\pdfoutput=1

\documentclass[11pt]{article}

\usepackage[preprint]{acl}

\usepackage{times}
\usepackage{latexsym}

\usepackage[T1]{fontenc}

\usepackage[utf8]{inputenc}

\usepackage{microtype}

\usepackage{inconsolata}

\usepackage{tikz}
\usetikzlibrary{shapes.geometric}
\usetikzlibrary{decorations.pathreplacing,calligraphy}
\usetikzlibrary {shapes.symbols} 
\usetikzlibrary{plotmarks}

\usepackage{sourcecodepro}

\usepackage{xcolor}
\definecolor{maroon}{cmyk}{0, 0.87, 0.68, 0.32}
\definecolor{halfgray}{gray}{0.55}
\definecolor{ipython_frame}{RGB}{207, 207, 207}
\definecolor{ipython_bg}{RGB}{247, 247, 247}
\definecolor{ipython_red}{RGB}{186, 33, 33}
\definecolor{ipython_green}{RGB}{0, 128, 0}
\definecolor{ipython_cyan}{RGB}{64, 128, 128}
\definecolor{ipython_purple}{RGB}{170, 34, 255}

\definecolor{lred}{RGB}{252, 224, 225}
\definecolor{lorange}{RGB}{255, 226, 187}
\definecolor{lyellow}{RGB}{253, 249, 192}
\definecolor{lblue}{RGB}{194, 232, 247}
\definecolor{lgreen}{RGB}{204, 231, 207}
\definecolor{lgrey}{RGB}{230, 230, 230}
\definecolor{lpurple}{RGB}{197, 190, 223}
\definecolor{lmagenta}{RGB}{243, 200, 220}
\definecolor{laqua}{RGB}{104, 165, 179}

\usepackage{listings}
\lstdefinelanguage{iPython}{
    morekeywords={access,and,break,class,continue,def,del,elif,else,except,exec,finally,for,from,global,if,import,in,is,lambda,not,or,pass,print,raise,return,try,while},%
    morekeywords=[2]{abs,all,any,basestring,bin,bool,bytearray,callable,chr,classmethod,cmp,compile,complex,delattr,dict,dir,divmod,enumerate,eval,execfile,file,filter,float,format,frozenset,getattr,globals,hasattr,hash,help,hex,id,input,int,isinstance,issubclass,iter,len,list,locals,long,map,max,memoryview,min,next,object,oct,open,ord,pow,property,range,raw_input,reduce,reload,repr,reversed,round,set,setattr,slice,sorted,staticmethod,str,sum,super,tuple,type,unichr,unicode,vars,xrange,zip,apply,buffer,coerce,intern},%
    sensitive=true,%
    morecomment=[l]\#,%
    morestring=[b]',%
    morestring=[b]",%
    morestring=[s]{'''}{'''},
    morestring=[s]{"""}{"""},
    morestring=[s]{r'}{'},
    morestring=[s]{r"}{"},%
    morestring=[s]{r'''}{'''},%
    morestring=[s]{r"""}{"""},%
    morestring=[s]{u'}{'},
    morestring=[s]{u"}{"},%
    morestring=[s]{u'''}{'''},%
    morestring=[s]{u"""}{"""},%
    literate=
    {á}{{\'a}}1 {é}{{\'e}}1 {í}{{\'i}}1 {ó}{{\'o}}1 {ú}{{\'u}}1
    {Á}{{\'A}}1 {É}{{\'E}}1 {Í}{{\'I}}1 {Ó}{{\'O}}1 {Ú}{{\'U}}1
    {à}{{\`a}}1 {è}{{\`e}}1 {ì}{{\`i}}1 {ò}{{\`o}}1 {ù}{{\`u}}1
    {À}{{\`A}}1 {È}{{\'E}}1 {Ì}{{\`I}}1 {Ò}{{\`O}}1 {Ù}{{\`U}}1
    {ä}{{\"a}}1 {ë}{{\"e}}1 {ï}{{\"i}}1 {ö}{{\"o}}1 {ü}{{\"u}}1
    {Ä}{{\"A}}1 {Ë}{{\"E}}1 {Ï}{{\"I}}1 {Ö}{{\"O}}1 {Ü}{{\"U}}1
    {â}{{\^a}}1 {ê}{{\^e}}1 {î}{{\^i}}1 {ô}{{\^o}}1 {û}{{\^u}}1
    {Â}{{\^A}}1 {Ê}{{\^E}}1 {Î}{{\^I}}1 {Ô}{{\^O}}1 {Û}{{\^U}}1
    {œ}{{\oe}}1 {Œ}{{\OE}}1 {æ}{{\ae}}1 {Æ}{{\AE}}1 {ß}{{\ss}}1
    {ç}{{\c c}}1 {Ç}{{\c C}}1 {ø}{{\o}}1 {å}{{\r a}}1 {Å}{{\r A}}1
    {€}{{\EUR}}1 {£}{{\pounds}}1
    {^}{{{\color{ipython_purple}\^{}}}}1
    {=}{{{\color{ipython_purple}=}}}1
    {+}{{{\color{ipython_purple}+}}}1
    {*}{{{\color{ipython_purple}$^\ast$}}}1
    {/}{{{\color{ipython_purple}/}}}1
    {+=}{{{+=}}}1
    {-=}{{{-=}}}1
    {*=}{{{$^\ast$=}}}1
    {/=}{{{/=}}}1,
    literate=
    *{-}{{{\color{ipython_purple}-}}}1
     {?}{{{\color{ipython_purple}?}}}1,
    identifierstyle=\color{black}\ttfamily,
    commentstyle=\color{ipython_cyan}\ttfamily,
    stringstyle=\color{ipython_red}\ttfamily,
    keepspaces=true,
    showspaces=false,
    showstringspaces=false,
    rulecolor=\color{ipython_frame},
    frame=single,
    frameround={t}{t}{t}{t},
    framexleftmargin=0mm,
    numberstyle=\tiny\color{white},
    backgroundcolor=\color{ipython_bg},
    basicstyle=\scriptsize,
    keywordstyle=\color{ipython_green}\ttfamily,
}

\lstdefinelanguage{iPython_output}{
    morekeywords={Output},%
    morekeywords=[2]{},%
    sensitive=true,%
    morecomment=[l]\#,%
    morestring=[b]',%
    morestring=[b]",%
    morestring=[s]{'''}{'''},
    morestring=[s]{"""}{"""},
    morestring=[s]{r'}{'},
    morestring=[s]{r"}{"},%
    morestring=[s]{r'''}{'''},%
    morestring=[s]{r"""}{"""},%
    morestring=[s]{u'}{'},
    morestring=[s]{u"}{"},%
    morestring=[s]{u'''}{'''},%
    morestring=[s]{u"""}{"""},%
    literate=
    {á}{{\'a}}1 {é}{{\'e}}1 {í}{{\'i}}1 {ó}{{\'o}}1 {ú}{{\'u}}1
    {Á}{{\'A}}1 {É}{{\'E}}1 {Í}{{\'I}}1 {Ó}{{\'O}}1 {Ú}{{\'U}}1
    {à}{{\`a}}1 {è}{{\`e}}1 {ì}{{\`i}}1 {ò}{{\`o}}1 {ù}{{\`u}}1
    {À}{{\`A}}1 {È}{{\'E}}1 {Ì}{{\`I}}1 {Ò}{{\`O}}1 {Ù}{{\`U}}1
    {ä}{{\"a}}1 {ë}{{\"e}}1 {ï}{{\"i}}1 {ö}{{\"o}}1 {ü}{{\"u}}1
    {Ä}{{\"A}}1 {Ë}{{\"E}}1 {Ï}{{\"I}}1 {Ö}{{\"O}}1 {Ü}{{\"U}}1
    {â}{{\^a}}1 {ê}{{\^e}}1 {î}{{\^i}}1 {ô}{{\^o}}1 {û}{{\^u}}1
    {Â}{{\^A}}1 {Ê}{{\^E}}1 {Î}{{\^I}}1 {Ô}{{\^O}}1 {Û}{{\^U}}1
    {œ}{{\oe}}1 {Œ}{{\OE}}1 {æ}{{\ae}}1 {Æ}{{\AE}}1 {ß}{{\ss}}1
    {ç}{{\c c}}1 {Ç}{{\c C}}1 {ø}{{\o}}1 {å}{{\r a}}1 {Å}{{\r A}}1
    {€}{{\EUR}}1 {£}{{\pounds}}1
    {^}{{{\color{ipython_purple}\^{}}}}1
    {=}{{{\color{ipython_purple}=}}}1
    {+}{{{\color{ipython_purple}+}}}1
    {*}{{{\color{ipython_purple}$^\ast$}}}1
    {/}{{{\color{ipython_purple}/}}}1
    {+=}{{{+=}}}1
    {-=}{{{-=}}}1
    {*=}{{{$^\ast$=}}}1
    {/=}{{{/=}}}1,
    literate=
    *{-}{{{\color{ipython_purple}-}}}1
     {?}{{{\color{ipython_purple}?}}}1,
    identifierstyle=\color{black}\ttfamily,
    commentstyle=\color{ipython_cyan}\ttfamily,
    stringstyle=\ttfamily,
    keepspaces=false,
    showspaces=false,
    showstringspaces=false,
    rulecolor=\color{ipython_frame},
    frame=single,
    frameround={t}{t}{t}{t},
    framexleftmargin=0mm,
    numberstyle=\tiny\color{white},
    backgroundcolor=\color{ipython_bg},
    basicstyle=\scriptsize,
    keywordstyle=\color{blue} \ttfamily,
}

\title{GPTopic: Dynamic and Interactive Topic Representations}

\author{
  \textbf{Arik Reuter\textsuperscript{1,2,3}},
  \textbf{Bishnu Khadka\textsuperscript{4, 5}},
  \textbf{Anton Thielmann\textsuperscript{3}},
    \textbf{Christoph Weisser\textsuperscript{4}},
    \\
  \textbf{Sebastian Fischer\textsuperscript{2,6}},
  \textbf{Benjamin S{\"a}fken\textsuperscript{3}},
\\
\\
    \textsuperscript{1}University of Cambridge,
  \textsuperscript{2}LMU Munich,
  \textsuperscript{3}TU Clausthal,
  \textsuperscript{4}BASF, \\
  \textsuperscript{5}Tribhuvan University,
  \textsuperscript{6}MCML
\\
  \small{
    \textbf{Correspondence:} \href{mailto:email@domain}{ar2364@cam.ac.uk}
  }
}

\begin{document}
\maketitle
\begin{abstract}
Topic modeling seems to be almost synonymous with generating lists of top words to represent topics within large text corpora. 
However, deducing a topic from such a list of individual terms can require substantial expertise and experience, making topic modelling less accessible to people unfamiliar with the particularities and pitfalls of top-word interpretation. A topic representation limited to top-words might further fall short of offering a comprehensive and easily accessible characterization of the various aspects, facets, and nuances a topic might have. To address these challenges, we introduce GPTopic, a software package that leverages Large Language Models (LLMs) to create dynamic, interactive topic representations. GPTopic provides an intuitive chat interface for users to explore, analyze, and refine topics interactively, making topic modeling more accessible and comprehensive.
The corresponding code is available here: \url{https://github.com/ArikReuter/TopicGPT}.

\end{abstract}

\section{Introduction}

Topic modeling has the goal of extracting topics from large corpora of text, thus revealing potentially highly interesting but otherwise latent common semantic themes. A topic itself is commonly represented as a list of “top-words” which comprise the terms best describing the identified topics \cite{blei2003latent, chemudugunta2006modeling, yin2014dirichlet, dieng2020topic}. For instance, the top-words “Elephant”, “Lion”, “Leopard”, “Buffalo”, “Rhinoceros” clearly imply the topic “African Animals”. Yet, in many real-world scenarios, outcomes are significantly less clear-cut and topics more nuanced. Therefore, deducing topics from a concise collection of terms demands both experience and expertise, which can be especially an issue in cases where scientists and practitioners apply topic modelling. We thus identify accessibility and interpretability as central issues in the subject of topic modelling, especially given that it is such a commonly used tool in numerous applications \cite{boyd2017applications, jelodar2019latent, thormann2021stock, kant2022iterative, tillmann2022privacy, thielmann2023unsupervised, weisser2023pseudo}.

While an increasing number of top words per topic would clearly cause a more detailed representation of a topic, it is questionable how many of those words humans can reasonably comprehend and interpret as a coherent theme. The top-word approach also carries an inherent risk of misinterpretation, especially when dealing with noisy data and abstract topics \cite{chang2009reading}. Additionally, from a practical perspective, a list of words is often simply too static to be useful for any interesting inquiries into the aspects, subtopics, and potentially complex nature of a topic. To put it in a nutshell: A topic is more than a list of words. 

We thus contribute the GPTopic software-package that utilizes Large Language Models (LLMs) to overcome the limited notion of top-word representations of topics in three ways:
\begin{enumerate}
    \item First, GPTopic allows to generate a concise name and a short description of topics, easily understandable by non-technical users. 
    
    \item Second, it further enables dynamic interactions with topics via a chat-based interface. This allows practitioners to extract more nuanced and specific information from topics. 

    \item Third, it is also possible to augment and modify a topic interactively based on the previous analyses. 
\end{enumerate}
In summary, we aim to make topic-modeling more interactive and more dynamic, but centrally more interpretable, open and especially more accessible. To accomplish enhanced accessibility and interpretability in a dynamic approach, we utilize LLMs to summarize and interact with topics. 

\section{Topic Extraction}

The core of the GPTopic package is built on top of a robust topic modeling approach that involves clustering document embeddings and then extracting the most relevant words using a method based on cosine similarity and a term frequency-inverse document frequency (TF-IDF) heuristic \cite{angelov2020top2vec, grootendorst2022bertopic}. By default, dimensionality reduction is performed using UMAP \cite{mcinnes2018umap}, and clustering is executed using the HDBSCAN algorithm \cite{mcinnes2017hdbscan}. 
Unlike \citet{angelov2020top2vec} and \citet{grootendorst2022bertopic}, we allow users to specify a fixed number of topics, where agglomerative clustering \cite{murtagh2014ward} is used to merge the clusters identified by HDBSCAN. 

Users have the flexibility to supply their own document embeddings or use embedding models from OpenAI, Google, or Sentence Transformers. The choice of embedding model can be aligned with the specific LLM selected, ensuring optimal compatibility and performance.


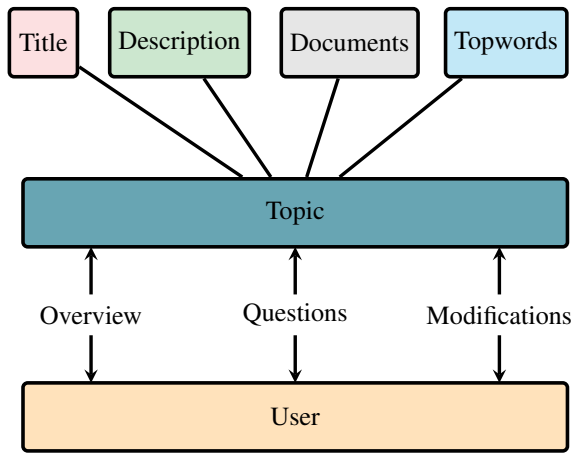
\begin{figure}[h]
    \centering
\scalebox{0.9}{
\begin{tikzpicture}

\node (Topic) [rectangle,rounded corners = 2pt, text centered, draw=black, fill=white, line width = 1.5pt, minimum width = 8cm, minimum height = 1cm, fill = laqua] {Topic};

\node (Title) [rectangle,rounded corners = 2pt, text centered, draw=black, fill=white, line width = 1.5pt, minimum width = 1cm, minimum height = 1cm, above of = Topic, yshift = 1.5cm, xshift = -3.7cm, fill = lred] {Title};
\node (Description) [rectangle,rounded corners = 2pt, text centered, draw=black, fill=white, line width = 1.5pt, minimum width = 1cm, minimum height = 1cm, above of = Topic, yshift = 1.5cm, xshift = -1.7cm, fill = lgreen] {Description};
\node (Documents) [rectangle,rounded corners = 2pt, text centered, draw=black, fill=white, line width = 1.5pt, minimum width = 1cm, minimum height = 1cm, above of = Topic, yshift = 1.5cm, xshift = 0.8cm, fill = lgrey] {Documents};
\node (Top-words) [rectangle,rounded corners = 2pt, text centered, draw=black, fill=white, line width = 1.5pt, minimum width = 1cm, minimum height = 1cm, above of = Topic, yshift = 1.5cm, xshift = 3.1cm, fill = lblue] {Topwords};

\node (User) [rectangle,rounded corners = 2pt, text centered, draw=black, fill=white, line width = 1.5pt, minimum width = 8cm, minimum height = 1cm, yshift = -2cm, below of = Topic, fill = lorange] {User};

\node (Overview) [rectangle, minimum width = 0cm, minimum height = 0cm, yshift = -0.5cm, xshift = -3cm, below of = Topic] {Overview};

\node (Questions) [rectangle,rounded corners = 0pt, text centered, minimum width = 0cm, minimum height = 0cm, yshift = -0.5cm, xshift = 0cm, below of = Topic] {Questions};

\node (Modifications) [rectangle,rounded corners = 0pt, yshift = -0.5cm, xshift = 3cm, below of = Topic] {Modifications};

\tikzstyle{arrow} = [->,>=stealth, line width = 0.5mm]
\tikzstyle{arrow_back} = [<-,>=stealth, line width = 0.5mm]
\tikzstyle{line} = [-,>=stealth, line width = 0.5mm]
\draw [line] (Topic) -- (Title);
\draw [line] (Topic) -- (Description);
\draw [line] (Topic) -- (Documents);
\draw [line] (Topic) -- (Top-words);

\draw [arrow_back] ([shift={(0,-1)}]Overview.center) -- ([shift={(0,-0.3)}]Overview.center);
\draw [arrow] ([shift={(0, 0.3)}]Overview.center) -- ([shift={(0,1)}]Overview.center);

\draw [arrow_back] ([shift={(0,-1)}]Questions.center) -- ([shift={(0,-0.3)}]Questions.center);
\draw [arrow] ([shift={(0, 0.3)}]Questions.center) -- ([shift={(0,1)}]Questions.center);

\draw [arrow_back] ([shift={(0,-1)}]Modifications.center) -- ([shift={(0,-0.3)}]Modifications.center);
\draw [arrow] ([shift={(0, 0.3)}]Modifications.center) -- ([shift={(0,1)}]Modifications.center);

\end{tikzpicture}
}

\caption{The GPTopic package allows a user to dynamically interact with a topic. A topic can be thought of as a structure defined by its documents, title, description, and top-words. Users cannot only read the topic's description but also ask questions and interactively modify the topic. Note that beyond what this figure shows, interactions on a more global level, e.g., comparisons of topics, are also possible. }

\end{figure}

\section{Topic Naming and Descriptions}

The process of generating names and descriptions for topics leverages the capabilities of either GPT \cite{openai2022chatgpt, bubeck2023sparks}, Gemini \cite{gemini2023}, or Claude \cite{bai2022constitutional}. Employing these Large Language Models (LLMs) facilitates the creation of intuitively interpretable and coherent topic representations in the form of natural text. This approach enables the utilization of extensive top-word sets comprising several hundred or even thousand words in order to accurately encapsulate the essence of each topic. By default, 500 top-words are used to extract a topic's title and description. In contrast, in standard practice, topics are typically characterized by a much more limited selection of ten to twenty top-words \cite{blei2003latent, dieng2020topic}.

\section{Interaction with Topics}
Treating topics as entities that are more than a static list of top-words can be achieved by allowing users to dynamically interact with the topics. For \mbox{GPTopic} the basis for this interaction is a set of topics, where each topic has its title, description, its assigned documents, and embeddings of the assigned documents.

\subsection{Questions}
In order to allow users to ask specific questions about a topic, we implement a Retrieval-Augmented-Generation (RAG) functionality \cite{lewis2020retrieval} that retrieves the most relevant documents with respect to the user's query from a specific topic. This retrieval step is implemented via embedding relevant keywords of the user's query followed by k-nearest neighbor search \cite{andoni2009nearest}.
Subsequently, an LLM is used to answer the user's question with the additional information.

Additionally, the GPTopic package provides a function that uses LLMs to identify the relevant topic for a given query, such that the topic index does not necessarily have to be known in advance. 

Further, we implement a function that provides an automatic comparison of topics based on their description. 

In summary, asking questions allows to easily access very specific and nuanced information a user might wish to acquire about one or several topics.

\begin{figure}[h]
    \centering
\scalebox{0.9}{
\begin{tikzpicture}

\node (LLM) [rectangle,rounded corners = 2pt, text centered, draw=black, fill=white, line width = 1.5pt, minimum width = 8cm, minimum height = 1cm, fill = lpurple] {Large Language Model};

\node (Prompt) [rectangle,rounded corners = 2pt, text centered, draw=black, fill=white, line width = 1.5pt, minimum width = 3cm, minimum height = 1cm, above of = Topic, yshift = 1cm, xshift = -2cm, fill = lyellow] {Prompt};
\node (Response) [rectangle,rounded corners = 2pt, text centered, draw=black, fill=white, line width = 1.5pt, minimum width = 3
cm, minimum height = 1cm, above of = Topic, yshift = 1cm, xshift = 2cm, fill = lred] {Response};

\node (Functions) [rectangle,rounded corners = 2pt, text centered, draw=black, fill=white, line width = 1.5pt, minimum width = 8cm, minimum height = 1cm, yshift = -1cm, xshift = 0cm, below of = Topic, fill = lgreen] {Functions};

\node (search) [rectangle,rounded corners = 2pt, text centered, draw=black, fill=white, line width = 1.5pt, minimum width = 1cm, text width = 1cm, minimum height = 1cm, yshift = -1cm, xshift = -3.35cm, below of = Functions, fill = laqua] {knn-search};

\node (compare) [rectangle,rounded corners = 2pt, text centered, draw=black, fill=white, line width = 1.5pt, minimum width = 1cm, text width = 2cm, minimum height = 1cm, yshift = -1cm, xshift = -1.25cm, below of = Functions, fill = lgrey] {Topic Comparison};

\node (etc) [rectangle, text centered, minimum width = 1cm, text width = 2cm, minimum height = 1cm, yshift = -1cm, xshift = 0.75cm, below of = Functions] {\large \ldots};

\node (split) [rectangle,rounded corners = 2pt, text centered, draw=black, fill=white, line width = 1.5pt, minimum width = 1cm, text width = 2cm, minimum height = 1cm, yshift = -1cm, xshift = 2.83cm, below of = Functions, fill = lblue] {Topic Splitting};

\tikzstyle{arrow} = [->,>=stealth, line width = 0.5mm]
\tikzstyle{arrow_back} = [<-,>=stealth, line width = 0.5mm]
\tikzstyle{arrow_double} = [<->,>=stealth, line width = 0.5mm]
\tikzstyle{line} = [-,>=stealth, line width = 0.5mm]

\draw [arrow] (Prompt) -- ([shift={(0,-1.5)}]Prompt.center);
\draw [arrow_back] (Response) -- ([shift={(0,-1.5)}]Response.center);

\draw [arrow] ([shift={(0,-2.5)}]Prompt.center) -- ([shift={(0,-3.5)}]Prompt.center);
\draw [arrow_back] ([shift={(0,-2.5)}]Response.center) -- ([shift={(0,-3.5)}]Response.center);

\draw [line] (Functions) -- ([shift={(0,0.5)}]search.center);
\draw [line] (Functions) -- ([shift={(0,0.5)}]compare.center);
\draw [line] (Functions) -- ([shift={(0,0.5)}]split.center);

\end{tikzpicture}
}

\caption{The chat-based interface for GPTopic is implemented by processing a user-defined prompt with an LLM. The LLM then decides which function to call. The result of this function call is processed with a further LLM-prompt and the final result is output.s. }

\end{figure}
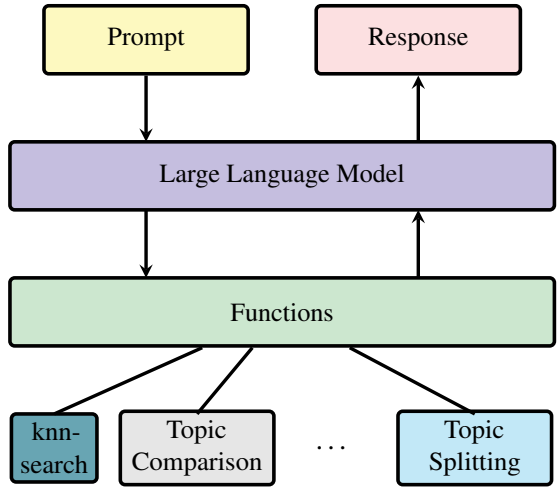

\subsection{Modifications}

After acquiring more specific information about a given topic modelling, it is  a natural feature to also adapt the topic modelling accordingly. Our software package provides several mechanisms to facilitate refinement of the initial topic structure. 

First, topics can be split in order to decrease the granularity of the topics. This is implemented in three different ways. We do not only provide topic splitting via running either k-means clustering \cite{lloyd1982least} or HDBSCAN clustering \cite{mcinnes2017hdbscan} on only the embeddings of a specified topic, but also allow splitting a topic based on a keyword. This keyword splitting works by embedding a user-defined keyword and then assigning all documents with embeddings closer to this new keyword than to the centroid of the embeddings of all documents of their original topic to a new topic.  Furthermore, a new topic based on a keyword can be created globally by assigning all documents to the new topic that are more similar to the keyword's embedding than the centroid of their topic's embeddings. The generation of new topics is always followed by top-word extraction, topic naming, and topic description of the modified topics. 

Second, the GPTopic package implements topic merging by simply combining the documents assigned to all sub-topics into a single super-topic, and subsequently recomputing the top-words, the topic's title and description. 

Finally, topics can also be deleted by assigning the documents of the deleted topics to the closest other topic. 

\subsection{A Chat-based Interface}

In order to make interacting with topics as easy as possible, the GPTopic package implements a chat-based interface. This works by processing prompts with an LLM-call that decides which function, among knn-search, topic splitting, topic combination, and all the other presented functions, to call with which parameters. The result of this function call is then combined with the original user-prompt and returned.

\section{Example Usage}
In this section, we demonstrate that GPTopic can be used very intuitively to extract topics and dynamically investigate them further. We use the Twenty Newsgroups corpus \cite{misc_twenty_newsgroups_113} to showcase the API our package provides. Note that for the purpose of representation, we pruned the outputs by the package and permuted the indices of the identified topics. 

Instantiating and fitting a model with the GPTopic package just takes a few lines of code. 

\begin{lstlisting}[language=iPython]
from gptopic.GPTopic import GPTopic

tm = TopicGPT(
    prompting_model=<your-choice-of-llm>,
    api_key=<your-provider-api-key>, #openai, google, anthropic
    n_topics=20,  # real number of topics
    embedding_model=<your-chosen-embedding-model>, 
    use_saved_embeddings=False # train from scratch
)
tm.fit(corpus)  # train the model on the corpus

\end{lstlisting}

\noindent As the next step, GPTopic allows access to the list of the names and descriptions of the topics, as extracted by the package
\begin{lstlisting}[language=iPython]
print(tm.topic_lis)
\end{lstlisting}
\begin{lstlisting}[language=iPython_output]
Output: 
[Topic 0: Electronics Equipment Sales,
 Topic 1: Space Exploration,
 Topic 2: Belief and Atheism
 ...
 Topic 19: Baseball games and teams]
\end{lstlisting}

\noindent We might be further interested in learning more about a detailed aspect of a topic and use a corresponding prompt to query the model: 

\begin{lstlisting}[language=iPython]
tm.pprompt("Which information related to the keyword 'moon landing' does topic 1 have?")
\end{lstlisting}

\begin{lstlisting}[language=iPython_output]
Output:
GPT wants to the call the function:  {
  "name": "knn_search",
  "arguments": "{'topic_index': 1, 'query': 'moon landing', 'k': 5}"
}
Topic 1, which is related to the keyword "moon 
landing," has the following information:

1. Document index 258: This document provides an 
introduction to the solar system and mentions that 
advancements in rocketry after World War II enabled 
machines to travel to the Moon and other planets. It 
highlights that the United States has sent both 
automated spacecraft and human-crewed expeditions to 
explore the Moon.

2. Document index 535: This document discusses a $65 
million program called the Back to the Moon bill, 
which aims to encourage private companies to develop 
lunar orbiters. 
[...]
\end{lstlisting}


\noindent The response of the model is a summary of relevant documents and how they relate to a specific query about the topic of interest. 

Another interesting point we want to further investigate could be the potential subtopics of \mbox{topic 2}: 
\begin{lstlisting}[language=iPython]
tm.pprompt("What are 5 potential subtopics of topic 2")
\end{lstlisting}

\begin{lstlisting}[language=iPython_output]
Output:
GPT wants to the call the function:  {
  "name": "split_topic_kmeans",
  "arguments": "{\n  \"topic_idx\": 2,\n  \"n_clusters\": 5\n}"
}

Here are five potential subtopics of topic 2:

1. Existence of God: This subtopic explores arguments, evidence, proofs, and the existence of God.

2. Atheism vs Theism: This subtopic delves into the concepts of atheism, theism, atheists, and theists, and the debates surrounding these belief systems.

[...]
\end{lstlisting}

\noindent The package thus decides to use a function for identifying five different subtopics within topic 2 and summarizes them. 

\section{Limitations and Shortcomings}

In our study, we observed that despite efforts to refine the prompt design, GPTopic may occasionally produce results that do not accurately reflect the content or themes of the documents it retrieves, which is caused by model hallucinations. Incorporating GPT-4 has shown promise in reducing hallucinations. 

We also note the importance of dataset size, recommending over 10,000 documents for optimal topic identification and the RAG mechanisms. 

Furthermore, while TopicGPT offers the benefits of modern LLMs, it forfeits the advantages of traditional Bayesian topic models that also output uncertainty over topic assignments \citep{blei2003latent, reuter2025probabilistic, weisser2023pseudo}.

\section{Related Work}

Several recent lines of work aim to make topic modelling more accessible. For instance the \mbox{OCTIS} package \cite{terragni2021octis} provides a comprehensive framework for training, analyzing, optimizing and comparing topic models. Gensim and Mallet are further well-established packages that implement topic-modelling functionalities \cite{rehurek_lrec, McCallum2002MALLET}. But neither package focuses on a final representation of topics beyond top-word lists. 

Further, the BERTopic package implements several intuitive topic visualizations, as well as the option to use LLMs for generating labels, summaries, phrases and keywords for existing topics \cite{grootendorst2022bertopic}. 

Very recent results by \citet{pham2023topicgpt} and \citet{wang2023prompting} explore the usage of prompting LLMs for topic extraction, and while they achieve strong benchmark results, they do not fundamentally challenge the notion of top-word based topic representations. 

Finally, \citet{sia2020tired} introduce a clustering-based topic modelling technique that employs weighting and re-ranking of top-words to achieve a more coherent topic representation. Similarly, an approach by \citet{thielmann2024topics}, implemented in the STREAM package \cite{thielmann2024stream}, questions the notion of static top-word based topic representations by introducing a method for increasing top-word coherence via using extension corpora. 

\section{Conclusion}

Throughout this paper, we have discussed the limitations inherent in traditional topic modeling approaches, which often rely on static lists of top-words to represent complex themes. These approaches, while foundational, can often fall short in capturing the dynamic and multifaceted nature of topics, especially in noisy or nuanced datasets. The introduction of the GPTopic package marks a step forward in addressing these challenges. By leveraging the capabilities of LLMs, GPTopic transcends the conventional boundaries of topic modeling, offering a more nuanced, interactive, and comprehensible method for topic analysis and interpretation. 

GPTopic's innovative approach, which enables the generation of concise topic names and descriptions, interactive engagement with topics, and the dynamic augmentation and refinement of topic representations, can be a pivotal tool for both researchers and practitioners.
It not only makes topic modeling more accessible to non-technical users but also enhances the interpretability and utility of topic modeling outcomes. As a result, GPTopic can facilitate a deeper understanding of text corpora for a broad community of people, enabling users to uncover and explore the underlying semantic themes in large corpora of text effectively.

\newpage



\bibliography{bib}

\appendix



\end{document}